%% file: irm_ite.tex
\documentclass{article}
\usepackage{spconf,amsmath,graphicx}
\usepackage{enumitem}
\usepackage{graphicx}
\usepackage{subfig,caption}
\usepackage{tikz}
\usetikzlibrary{arrows}
\usepackage{hyperref}
\usepackage{xcolor}

\ninept
\usepackage{amssymb}

\newcommand{\textBlue}[1]{{\leavevmode\color{blue}#1}} 
\newcommand{\textRed}[1]{{\leavevmode\color{red}#1}} 

\DeclareMathAlphabet{\mathbsf}{OT1}{cmss}{bx}{n}
\DeclareMathAlphabet{\mathssf}{OT1}{cmss}{m}{sl}
\newcommand{\rvx}{{\mathssf{x}}}	
\newcommand{\rvy}{{\mathssf{y}}}	
\newcommand{\rvt}{{\mathssf{t}}}	
\newcommand{\rve}{{\mathssf{e}}}	
\newcommand{\rvbx}{{\mathbsf{x}}} 
\newcommand{\svbx}{{\mathbf{x}}} 
\newcommand{\bQ}{{\mathbf{Q}}} 
\newcommand{\cX}{\mathcal{X}} 
\newcommand{\cY}{\mathcal{Y}} 
\newcommand{\cD}{\mathcal{D}} 
\newcommand{\cN}{\mathcal{N}} 
\newcommand{\cE}{\mathcal{E}} 
\newcommand{\cP}{\mathcal{P}} 
\newcommand{\Ite}{\tau} 
\newcommand{\estimatedIte}{\hat{\tau}} 
\newcommand{\Expectation}{\mathbb{E}} 
\newcommand{\pehe}{\epsilon_{PEHE}} 
\newcommand{\bmu}{\boldsymbol{\mu}} 
\newcommand{\bSigma}{\boldsymbol{\Sigma}} 
\newcommand{\bLambda}{\boldsymbol{\lambda}} 
\newcommand{\numEnv}{n_e} 

\def\rad{4pt}
\newcommand*\blueVerNoSpace{{\tikz [baseline=-0.8*\rad, color= blue]{
    \draw  [clip](0,-\rad) arc [start angle=270, delta angle=180, radius=\rad];
    \draw [blue] (0,-\rad) arc [start angle=270, delta angle=180, radius=\rad];
    \draw [blue,line width=0.25mm] (0, -\rad) -- (0,\rad);
    \draw [blue] (\rad/4, -\rad) -- (\rad/4,\rad);
    \draw [blue] (2*\rad/4, -\rad) -- (2*\rad/4,\rad);
    \draw [blue] (3*\rad/4, -\rad) -- (3*\rad/4,\rad);
}}}

\newcommand*\blueHorNoSpace{{\tikz [baseline=-0.8*\rad, color= blue]{
    \draw  [clip](0,\rad) arc [start angle=90, delta angle=180, radius=\rad];
    \draw [blue] (0,\rad) arc [start angle=90, delta angle=180, radius=\rad];
    \draw [blue,line width=0.25mm] (0, \rad) -- (0,-\rad);
    \draw [blue] (-\rad, 0) -- (0, 0);
    \draw [blue] (-\rad, \rad/4) -- (0, \rad/4);
    \draw [blue] (-\rad, -\rad/4) -- (0, -\rad/4);
    \draw [blue] (-\rad, 2*\rad/4) -- (0, 2*\rad/4);
    \draw [blue] (-\rad, -2*\rad/4) -- (0, -2*\rad/4);
    \draw [blue] (-\rad, 3*\rad/4) -- (0, 3*\rad/4);
    \draw [blue] (-\rad, -3*\rad/4) -- (0, -3*\rad/4);
}}}

\newcommand*\redCircleNoSpace{{\tikz [baseline=-0.8*\rad]{
   \draw  [red!90](0,0) circle (\rad);
}}}

\newcommand*\blueCircleNoSpace{{\tikz [baseline=-0.8*\rad]{
   \draw  [blue](0,0) circle (\rad);
}}}

\newcommand*\blackCircleNoSpace{{\tikz [baseline=-0.8*\rad]{
   \draw  [black](0,0) circle (\rad);
}}}

\newcommand*\redCircleDashedNoSpace{{\tikz [baseline=-0.8*\rad]{
   \draw  [red!90,densely dashed](0,0) circle (\rad);
}}}

\newcommand*\blueCircleDashedNoSpace{{\tikz [baseline=-0.8*\rad]{
   \draw  [blue,densely dashed](0,0) circle (\rad);
}}}

\newcommand*\blackCircleDashedNoSpace{{\tikz [baseline=-0.8*\rad]{
   \draw  [black,densely dashed](0,0) circle (\rad);
}}}

\newcommand*\centerLineNoSpace{{\tikz [baseline=-0.8*\rad]{
 	\draw [gray, line width=0.25mm, -] (-0.75*\rad, 0.75*\rad) -- (0.75*\rad,-0.75*\rad); 
	 \draw  [gray, fill=gray](0.75*\rad,-0.75*\rad) circle (0.02cm); 
	 \draw  [gray, fill=gray](-0.75*\rad,0.75*\rad) circle (0.02cm); 
 }}}
 \newcommand{\centerLine}{\centerLineNoSpace$\hspace{1mm}$}
 
 \newcommand{\redCircle}{\redCircleNoSpace$\hspace{1mm}$}
  \newcommand{\blueCircle}{\blueCircleNoSpace$\hspace{1mm}$}
 \newcommand{\blackCircle}{\blackCircleNoSpace$\hspace{1mm}$}
 
 \newcommand{\blueCircleDashed}{\blueCircleDashedNoSpace$\hspace{1mm}$}
  \newcommand{\redCircleDashed}{\redCircleDashedNoSpace$\hspace{1mm}$}
   \newcommand{\blackCircleDashed}{\blackCircleDashedNoSpace$\hspace{1mm}$}

 \newcommand{\blueVer}{\blueVerNoSpace$\hspace{1mm}$}

\title{TREATMENT EFFECT ESTIMATION USING INVARIANT RISK MINIMIZATION}
\name{Abhin Shah,$^{\dagger,\star}$ Kartik Ahuja,$^{\dagger}$ Karthikeyan Shanmugam,$^{\dagger}$ Dennis Wei,$^{\dagger}$}
\secondlinename{Kush R.~Varshney,$^{\dagger}$ and Amit Dhurandhar$^{\dagger}$}
\address{$^{\dagger}$IBM Research, $^{\star}$Massachusetts Institute of Technology}
\begin{document}
%
\maketitle
  \begin{abstract}
Inferring causal individual treatment effect (ITE) from observational data is a challenging problem whose difficulty is exacerbated by the presence of treatment assignment bias.  
In this work, we propose a new way to estimate the ITE using the domain generalization framework of invariant risk minimization (IRM).
IRM uses data from multiple domains, learns predictors that do not exploit spurious domain-dependent factors, and generalizes better to unseen domains.
We propose an IRM-based ITE estimator aimed at tackling treatment assignment bias when there is little support overlap between the control group and the treatment group.
We accomplish this by creating \textit{diversity}:  given a single dataset, we split the data into multiple domains artificially.
These diverse domains are then exploited by IRM to more effectively generalize 
regression-based models to data regions that lack support overlap. 
We show gains over classical regression approaches to ITE estimation in settings when support mismatch is more pronounced.
\end{abstract}
\begin{keywords}
Causal inference, individual treatment effect 

estimation, invariant risk minimization
\end{keywords}
\section{Introduction}
\label{sec_introduction}
\input{tex/introduction}
\input{tex/related_work}
\input{tex/contributions}
\input{tex/example}
\input{tex/problem_formulation}
\input{tex/approach}

\input{tex/experiments}
\input{tex/conclusion}

\clearpage
\bibliographystyle{IEEEbib}
\bibliography{refs}

\end{document}

%% file: tex/introduction.tex
Estimating the individual-level causal effect of a treatment is a fundamental problem in causal inference and applies to many fields. A few examples include understanding how a certain medication affects a patient's health \cite{Shalit2017, Alaa2017}, understanding how Yelp ratings influence a potential restaurant customer \cite{Anderson2012}, estimating the influence of individuals in social networks \cite{Smith2018}, inferring the effect of a policy in recommendation systems \cite{Schnabel2016}, assessing the causal impact of the treatment reception in sensor networks \cite{Coates2004}, estimating the impact of demand response signals \cite{Li2016}, and evaluating the effect of a policy on unemployment rates \cite{Lalonde1986}. Traditionally, randomized control trials (RCTs) have been used to evaluate treatment effects, but they can often be expensive and in some cases unethical.

In most scenarios, observational data that contains past actions and their responses is readily available. However,  observational data does not provide access to the causal reasoning behind a particular action. See Table \ref{table:1} for an illustrative example of a hospital record where age and blood pressure are features of patients, and blood sugar (either `low' or `high') is the response to a drug (either `0' or `1'). For a binary treatment, one of the options is often referred to as \textit{the control} (say drug `0') and the other one as \textit{the treatment} (say drug `1'). The group of individuals receiving \textit{the control} is collectively referred to as \textit{the control group} (patients `A', `B' in Table \ref{table:1}), and the group of individuals receiving \textit{the treatment} is collectively referred to as \textit{the treatment group} (patients `C', `D', `E' in Table \ref{table:1}).

The individual treatment effect (ITE) of a binary treatment is the difference between the outcome under \textit{the treatment} and the outcome under \textit{the control}. Estimating ITE from observational data differs from classical supervised learning because we never observe the ITE in our training data. For example,  in Table \ref{table:1} we do not observe the blood sugar under \textit{the treatment} for patients in the \textit{control group} and the blood sugar under \textit{the control} for patients in the \textit{treatment group}.

Unlike RCTs, observational data is often prone to treatment assignment bias \cite{Rosenbaum2002}. For instance, patients receiving drug `0' may have a higher natural tendency (due to their age) to have low blood sugar than patients receiving drug `1'. In other words, sub-populations receiving different treatments can have very different distributions, and a traditional supervised learning  model trained to predict the effect of treatment would fail to generalize well to the entire population. This issue calls 
for domain generalization methods for ITE estimation; in this paper, we make progress in this direction.
\begin{table}[t]
	\centering
	\caption{A typical observational record from a hospital} 
{
	\begin{tabular}{c|c|c|c|c}
	\hline
		Patient&Age&Blood Pressure&Drug&Blood sugar\\
		\hline
		\hline
		A & 22 & 145/95 & 0 & Low\\
		B & 26 & 135/80 & 0 & Low\\
		C & 58 & 130/70 & 1 & Low\\
		D & 50 & 145/80 & 1 & High\\
		E & 24 & 150/85 & 1 & Low\\
	\end{tabular}}
	\label{table:1}
	\vspace{-.1in}
\end{table}

%% file: tex/related_work.tex
\vspace{-1mm}
\subsection{Related Works}
\label{subsec_related_works}
\textbf{Covariate adjustment.} Existing works on covariate adjustment, a popular approach in treatment effect estimation, can be divided into two broad categories (a) balancing/matching, and (b) regression adjustment.
Classical techniques for balancing rely on propensity score estimation \cite{Rosenbaum1983}.
Propensity score weighting \cite{Austin2011} re-weights the samples to make \textit{the treatment group} and \textit{the control group} more similar. 
A few approaches \cite{Kallus2017, Gretton2009, Kallus2018} directly minimize imbalance metrics like kernel maximum mean discrepancy or discriminative discrepancy. 
Classical matching techniques match the samples from \textit{the treatment group} and \textit{the control group} using nearest neighbor matching \cite{Rubin1973, Abadie2004} or optimal matching \cite{Rosenbaum1989}. More recent methods match using the estimated propensity score \cite{Austin2011}, coarsened versions of the observed covariates \cite{Iacus2012}, or cardinality matching \cite{Visconti2018}. 

Regression adjustment estimates the potential outcomes with a supervised learning model fit on the features and the treatment. There are two main categories: (a) the T-learner (T for `two') that uses separate base-learners to estimate the outcome under \textit{control} and under \textit{treatment} and
(b) the S-learner (S for `single') that uses one base-learner to estimate the outcome using the features and the treatment assignment, without giving the treatment assignment any special role. This terminology comes from \cite{Kunzel2019}, and we will use it throughout this paper. Ordinary least squares (OLS) regression, which solves the empirical risk minimization (ERM) problem for square loss and linear function class, is one traditional choice as the base-learner for T-learner and S-learner. We denote these  by OLS/LR2 and OLS/LR1, respectively. Advanced machine learning (ML) methods such as tree-based models \cite{Wager2018, Athey2016, Hill2011} and deep generative models like generative adversarial networks, variational autoencoders, and multi-task Gaussian processes \cite{Alaa2017, Yoon2018, Louizos2017} have also been employed. 

\textbf{Domain Adaptation and Generalization.} In domain adaptation for supervised learning, the learner exploits the access to labeled data from the training domain and unlabeled data from test domain and performs well on the test domain.  In domain adaptation-driven ITE estimation methods \cite{Shalit2017}, the labeled training data consists of outcomes under \textit{the treatment} of \textit{the treatment group} and unlabeled test data is \textit{the control group} for which \textit{the treatment} outcomes are unknown. Recent works inspired by domain adaptation \cite{Shalit2017, Johansson2016, Shi2019} focus on learning new feature representations using neural architectures to match \textit{the treatment group} and \textit{the control group} in the representation space.  This is effective when the learned feature representation is strongly ignorable (no unmeasured confounding \cite{Imbens2009}). However, the usual strong ignorability assumption might not hold for this learned representation even if it holds for the original features. Domain generalization methods \cite{matsuura2020domain,gulrajani2020search,Arjovsky2019} in supervised learning use labeled data from multiple training domains while not requiring any unlabeled test data and learn models that generalize well to unseen domains. Domain generalization based methods seem to offer several advantages over domain adaptation in supervised learning but have not been explored for ITE estimation, which is the objective of this work. In our work, we rely on a recent domain generalization framework called invariant risk minimization (IRM) \cite{Arjovsky2019}. 
The IRM uses the following principle to perform well on unseen domains: rely on features whose predictive power is invariant across domains,  and ignore features whose predictive power varies across domains.
Note that our usage of IRM for ITE  does not rely on any additional ignorability assumptions on intermediate representations learned. 

%% file: tex/contributions.tex
\vspace{-1.5mm}
\subsection{Contributions}
\vspace{-0.5mm}
\label{subsec_contributions}
In this work, we explore the idea of domain generalization for ITE estimation from observational data. More specifically, we propose a new way to estimate ITE by bridging the framework of IRM and causal effect estimation. Our estimator is most effective when there is limited overlap in the support between \textit{the control group} and \textit{the treatment group}. Although the data comes from a single domain, we artificially create the diverse domains required for IRM. 
We provide an intuitive explanation of how IRM uses these diverse domains to tackle treatment assignment bias when there is little support overlap.
We support this with experiments and show gains over OLS/LR1 (linear S-learner) and OLS/LR2 (linear T-learner) in various settings when support mismatch is more pronounced. 

\textbf{Comparisons.} 
For a first evaluation of IRM in ITE estimation, we consider experiments in a simpler linear setting with the necessary interaction term for heterogeneity. Our primary approach uses the IRM framework as the base-learner for the T-learner (denoted by IRM$_2$) and is most comparable to OLS/LR2. The base-learner of OLS/LR2 that estimates the outcome under \textit{the control} does not use any information about the feature distribution of \textit{the treatment group}, similar to IRM$_2$. This is in contrast to OLS/LR1 that uses the feature distribution of both \textit{the control group} and \textit{the treatment group} to estimate the outcome under \textit{the control}. For the sake of completeness, we also use the IRM framework as the base-learner for the S-learner (denoted by IRM$_1$) and compare with both OLS/LR2 and OLS/LR1. We defer the comparison of our approach with ITE estimation approaches that use non-linear ML methods for future work.


%% file: tex/example.tex
\section{A toy example}
\label{sec_example}
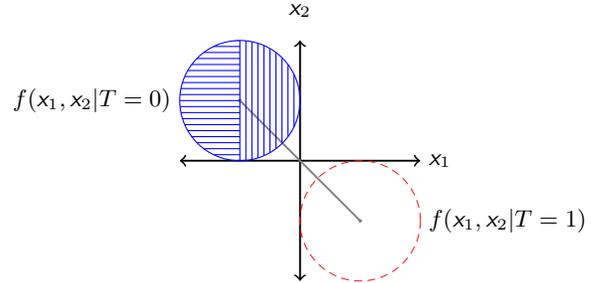
\begin{figure}
\centering
{\begin{tikzpicture}[scale=0.8]
\centering
    \begin{scope}[color=blue]
    	 \draw [black, line width=0.25mm, <->] (1, -3) --(1,1); 
	 \draw [black, line width=0.25mm, <->] (-1, -1) -- (3,-1); 
	 \draw [gray, line width=0.25mm, -] (0, 0) -- (2,-2); 
	 \draw  [gray, fill=gray](2,-2) circle (0.02cm); 
	 \draw  [gray, fill=gray](0,0) circle (0.02cm); 
    	\draw  [red!90, densely dashed](2,-2) circle (1cm); 
        \draw  [clip](0,0) circle (1cm); 
        \draw [blue] (0,-1) -- (0,1); 
        \draw[blue]  (-1, 0) -- (0,0); 
            \foreach \x in {1,...,10} {
                \draw [blue] (\x/10, -1) -- (\x/10,1); 
                \draw [blue] (-1, \x/10) -- (0,\x/10); 
                \draw [blue] (-1, -\x/10) -- (0,-\x/10); 
}
    \end{scope}
    \node[right] at (3,-1){$\rvx_1$};
    \node[above] at (1,1.25){$\rvx_2$};
    \node[left] at (-1,0){$f(\rvx_1, \rvx_2|T = 0)$};
    \node[right] at (3,-2){$f(\rvx_1, \rvx_2|T = 1)$};
\end{tikzpicture}}
\caption{A toy example of an observational data to illustrate the intuition behind applying the IRM framework to ITE estimation} \label{fig:illustrative_example}
\end{figure}

 Consider the illustrative example  in Figure \ref{fig:illustrative_example} with a binary treatment $T$. The feature distribution for \textit{the control group} ($f(\rvx_1, \rvx_2|T = 0)$) is in \textBlue{blue} and the feature distribution for \textit{the treatment group} ($f(\rvx_1, \rvx_2|T = 1)$) is in \textRed{red}. As shown, this is a case of a support mismatch between the two groups. We  use a full circle i.e., \blackCircle and a dashed circle i.e., \blackCircleDashed to denote the treatment assignment $T = 0$ and $T = 1$ respectively. Given observational data, we have access to the outcome under $T = 0$ for the \textit{control group} i.e., \blueCircle and the outcome under $T = 1$ for the \textit{treatment group} i.e., \redCircleDashedNoSpace. We aim to estimate the outcome under $T = 1$ for the \textit{control group} i.e., \blueCircleDashed and the outcome under $T = 0$ for the \textit{treatment group} i.e., \redCircleNoSpace. 

Let us first focus on the T-learner. Our first base-learner (say \textit{the control} branch) is supposed to learn the outcome for $T = 0$ using only \blueCircle (i.e., training data) and estimate the outcome for \redCircle (i.e., test data). Similarly, our second base-learner (say \textit{the treatment} branch) is supposed to learn the outcome for $T = 1$ using only \redCircleDashed (i.e., training data) to estimate the outcome for \blueCircleDashed (i.e., test data). Each of these base-learners is required to do domain generalization to 

Let us look at \textit{the control} branch in detail. If we were to use OLS as the base-learner, then with access to finite data, it will pick up the spurious correlations (induced by treatment assignment bias) in \blueCircle and fail to generalize well. In other words, the OLS/LR2 trained on \blueCircle will perform well on individuals from \blueCircle but will fail to do well on individuals from \redCircleNoSpace. If we were to use the IRM as the base-learner, we first need to split \blueCircle into multiple domains (say \blueVer and \blueHorNoSpace) so as to have varying levels of spurious correlations in \blueVer and \blueHorNoSpace. By training on \blueVer and \blueHorNoSpace, \textit{the control} branch of IRM$_2$ learns how to transport between \blueVer and \blueHorNoSpace. Being a domain generalization method, we expect the IRM method to generalize well on \redCircle which is outside the convex hull of the training data (i.e., outside \blueCircleNoSpace). As the support overlap  between \textit{the control group} and \textit{the treatment group} increases, we will see in Section \ref{sec_experiments} that gains of IRM$_2$ over OLS/LR2 and OLS/LR1 become more prominent.

Let us now focus on the S-learner that uses a single base-learner to learn the outcomes for both $T = 0$ and $T = 1$ using \blueCircle and \redCircleDashed (i.e., training data) to estimate the outcome for \blueCircleDashed and \redCircle (i.e., test data). As before, OLS will pick spurious correlations and fail to generalize well, but we still expect the IRM to generalize well. However, this domain generalization is not as straightforward as the T-learner because the treatment assignment is treated in a similar fashion as the other features of an individual, and there is lesser information for IRM to exploit the invariant factors across the domains. 


%

%% file: tex/problem_formulation.tex
\section{Problem Formulation}
\label{sec_problem_formulation}
\subsection{Setup}
We adopt the Rubin-Neyman potential outcomes framework \cite{Rubin1974}.

\noindent Let $\cX$ be the $d$-dimensional feature space, $\cY$ be the outcome space, and $\rvbx \in \cX$ be the $d$-dimensional feature vector. 
Let $\rvt = \{0,1\}$ be the binary treatment variable with $\rvt = 1$ being \textit{the treatment} and $\rvt = 0$ being \textit{the control}.
For $i \in \{0,1\}$, let $\rvy_i \in \cY$ be the potential outcome under $t = i$. Let $\svbx \in \cX, t \in \{0,1\}, y_0 \in \cY, y_1 \in \cY$ denote realizations of $\rvbx, \rvt, \rvy_0, \rvy_1$ respectively.
Suppose we have an observational dataset of $n$ individuals where for each individual we only observe the potential outcome that corresponds to the assigned treatment (denoted by $y_f$ and referred to as the \textit{factual outcomes}). 
Let our dataset be $\cD^{(n)} = \{\svbx^{(i)},t^{(i)},y_f^{(i)}\}_{i=1}^{n}$ where $y_f^{(i)} = y_0^{(i)}$ if $t^{(i)} = 0$ and $y_f^{(i)} = y_1^{(i)}$ if $t^{(i)} = 1$. Let $[n] = \{1,\cdots,n\}$.

We assume the standard strong ignorability condition: $0< p(\rvt = 1 | \rvbx) < 1$ and $(\rvy_0,\rvy_1) \perp \!\!\! \perp \rvt | \rvbx$ for all $\rvbx$. This is a sufficient condition for ITE to be identifiable from observational data \cite{Imbens2009, Pearl2017}. 

\vspace{-1mm}
\subsection{Inference Tasks}
Our interest lies in learning the ITE (denoted by $\Ite$) defined as:
\begin{align}\label{ite}
\Ite^{(i)} = y_1^{(i)} - y_0^{(i)} \hspace{3mm} \mbox{$\forall i \in [n]$}
\end{align}
Empirically, the estimate $\estimatedIte$ is evaluated using the \textit{precision in estimation of heterogeneous effect} (PEHE) which is the mean squared error of the estimated ITE for all the individuals in our data:
\begin{align}\label{pehe}
\pehe = \frac{1}{n} \sum_{i=1}^n (\Ite^{(i)}  - \estimatedIte^{(i)} )^2
\end{align}
\vspace{-3mm}
\subsection{Invariant Risk Minimization}
\cite{Arjovsky2019} consider datasets $D_{e}$, consisting of observations of the feature vector ($\rvbx \in \cX$) and the response ($\rvy \in \cY$), collected under multiple training domains $e \in \cE_{tr}$.
The dataset $D_{e} $, from domain $e$, contains i.i.d.\ samples according to some probability distribution $\cP_e$. The goal is to use these multiple datasets to learn a predictor $f: \cX \to \cY$ that minimizes the maximum risk over all the  domains $\cE$ i.e., $\min_f \max_{e \in \cE} R_e(f)$ where $R_e(f) = \Expectation_{(\rvbx, \rvy)\sim \cP_e} [l(f(\rvbx), \rvy)]$ is the risk under domain $e$ for a convex and differentiable loss function $l$. The practical version of IRM (i.e., IRMv1) is as follows:
\begin{align}\label{irmv1}
    \min_{\Phi: \cX \rightarrow \cY}\quad &\sum_{e \in \cE_{tr}} R_e(\Phi) + \lambda \| \nabla_{w | w = 1.0} R_e(w\cdot \Phi)\|^2
\end{align}
where $\Phi: \cX \rightarrow \cY$ is an invariant predictor, $w = 1.0$ is a fixed ``dummy'' classifier, the gradient norm penalty measures the optimality of the dummy classifier at each domain $e$. The first term in \eqref{irmv1} is a standard ERM term and $\lambda \in [0,\infty)$ is a regularizer balancing between this, and the invariance of the predictor $1 \cdot \Phi(\svbx)$. \cite{Arjovsky2019} solves IRMv1 in \eqref{irmv1} using stochastic gradient descent (SGD).


%% file: tex/approach.tex
\section{Our approach}
\label{sec:our_approach}
We do not assume access to multiple domains as required by IRM. Given access to a dataset from a single domain, we first split the dataset into different components representing diverse domains. The next step is the application of IRM.

\vspace{-1.5mm}
\subsection{Domain Generation}
\label{subsec_envgen}
We split $\cD^{(n)}$ into $\numEnv$ components as if each component is obtained from a different domain. To achieve this, we assign a variable $\rve \in [\numEnv]$ to each individual denoting which domain we place it in. We have $\cD^{(n)} = \cup_{j = 1}^{\numEnv} \cD^{(n)}_j$ where $\cD^{(n)}_{j} = \{(\svbx^{(i)}, t^{(i)}, y_f^{(i)}) : i \in [n], e^{(i)} = j\}$. We explored a variety of domain generation schemes. It turns out that, for our relatively simple setup, (uniformly) random domain generation is sufficient\footnote{For our setup, with relatively little data ($200$ training samples) and in high dimensions, the random scheme is sufficient as the domains appear sufficiently different to the different learners. We do not claim that the random scheme would work all the time.} i.e., $\rve$ takes any value in $[n_e]$ with the same probability.
\subsection{Procedure}
\label{subsec_procedure}
Let there be $n_{tr}$ training samples. Let $\cD^{(n_{tr})}_j$ be the component of $\cD^{(n_{tr})}$ corresponding to the $j^{th}$ domain. Let $\cD^{(n_{te})} = \{\svbx^{(i)}\}_{i=1}^{n_{te}}$ denote the test dataset consisting of $n_{te}$ samples.

\begin{enumerate}[leftmargin=*]
\item \textbf{T-learner / IRM$_2$}: 
For $j \in [n_e]$, let $\cD^{(n_{tr})}_{j,c} = \{(\svbx^{(i)}, y_f^{(i)}) : i \in [n_{tr}], t^{(i)} = 0, e^{(i)} = j\}$ be the \textit{control} component of $\cD^{(n_{tr})}_j$. Similarly, let $\cD^{(n_{tr})}_{j,t} = \{(\svbx^{(i)}, y_f^{(i)}) : i \in [n_{tr}], t^{(i)} = 1, e^{(i)} = j\}$ be the \textit{treatment} component of $\cD^{(n_{tr})}_j$.
\begin{itemize}[leftmargin=*]
\item \textbf{Training}. $\cD^{(n_{tr})}_{j,c}$ $\forall j \in [n_e]$ is training data for \textit{the control} branch of IRM$_2$.
$\cD^{(n_{tr})}_{j,t}$ $\forall j \in [n_e]$ is training data for \textit{the treatment} branch of IRM$_2$. Following \cite{Arjovsky2019}, we use SGD to optimize IRMv1 in \eqref{irmv1} for both branches of IRM$_2$.
\item \textbf{Testing}. Predict \textit{the control} outcomes on $\cD^{(n_{te})}$ using \textit{the control} branch of IRM$_2$ and  \textit{the treatment} outcomes on $\cD^{(n_{te})}$ using \textit{the treatment} branch of IRM$_2$.
\end{itemize}

\item \textbf{S-learner / IRM$_1$}: 
For $j \in [n_e]$, let $\hat{\cD}^{(n_{tr})}_{j} = \{(\svbx^{(i)}, t^{(i)}, \svbx^{(i)} \times t^{(i)}, y_f^{(i)}) : i \in [n], e^{(i)} = j\}$ where $\svbx \times t$ is the interaction term.
\begin{itemize}[leftmargin=*]
\item \textbf{Training}. $\hat{\cD}^{(n_{tr})}_{j}$ $\forall j \in [n_e]$ is the training data for the IRM$_1$. Following \cite{Arjovsky2019}, we use SGD to optimize IRMv1 in \eqref{irmv1} for IRM$_1$.
\item \textbf{Testing}. Using the trained IRM$_1$ framework, predict \textit{the control} outcome on $\cD^{(n_{te})}_{c} = \{(\svbx^{(i)},0,0)\}_{i=1}^{n_{te}}$ and \textit{the treatment} outcome on $\cD^{(n_{te})}_{t} = \{(\svbx^{(i)},1,\svbx^{(i)})\}_{i=1}^{n_{te}}$.
\end{itemize}
\end{enumerate}
OLS/LR1 and OLS/LR2 can be understood as unpenalized cases ($\lambda = 0$) of \eqref{irmv1} and with $n_e = 1$ in the above procedure.

%% file: tex/experiments.tex
\section{Experiments}
\label{sec_experiments}
\input{tex/data_generation}
\subsection{ITE estimation}
We consider 4 data generation schemes: (a) model A with linear outcome, (b) model B with linear outcome, (c) model A with quadratic outcome, (d) model B with quadratic outcome. We consider $n_{tr} = 200$ train samples, $n_{te} = 100$ test samples, and $n_{e} = 3$ domains. We average our results over 10 repetitions.
\begin{figure}[t]
  \centering
  \subfloat{\includegraphics[scale=0.25]{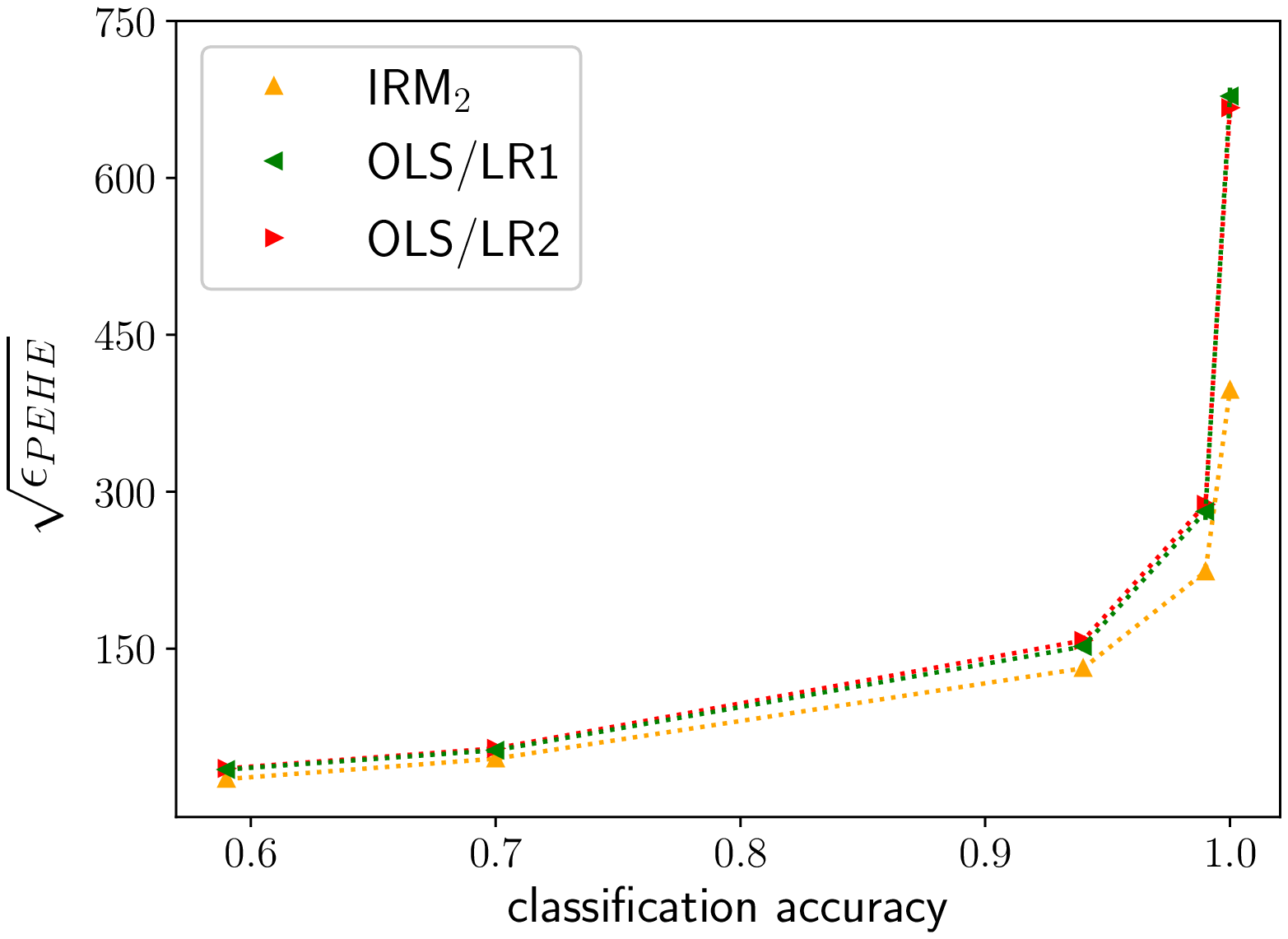}}
  \quad
  \subfloat{\includegraphics[scale=0.25]{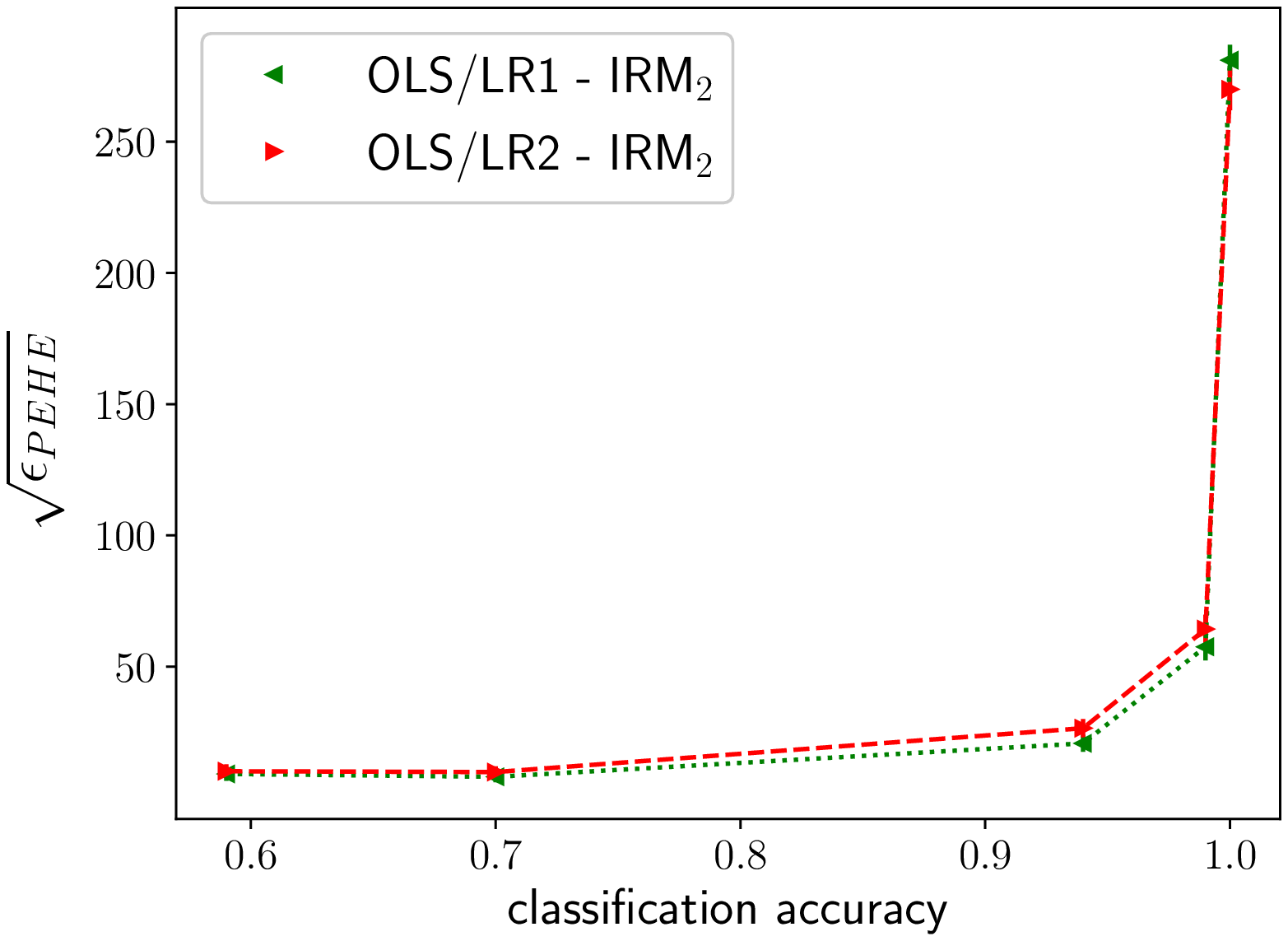}}
  \caption{$\sqrt{\pehe}$ (left) and $\sqrt{\pehe}$ difference (right) vs treatment group classification accuracy for model A with quadratic outcomes}
  \label{fig:class_A}
\end{figure}

\begin{figure}[t]
  \centering
  \subfloat{\includegraphics[scale=0.25]{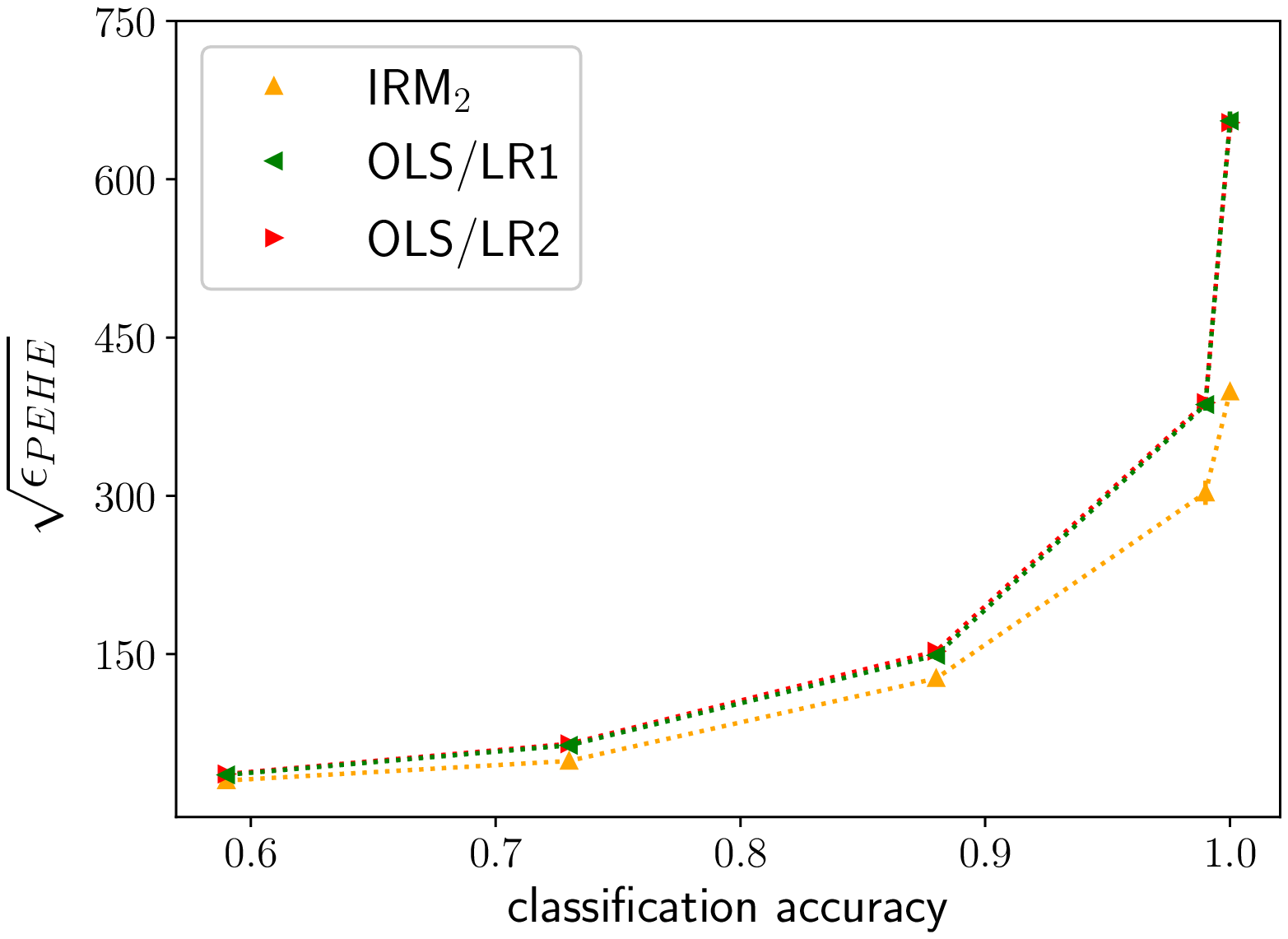}}
  \quad
  \subfloat{\includegraphics[scale=0.25]{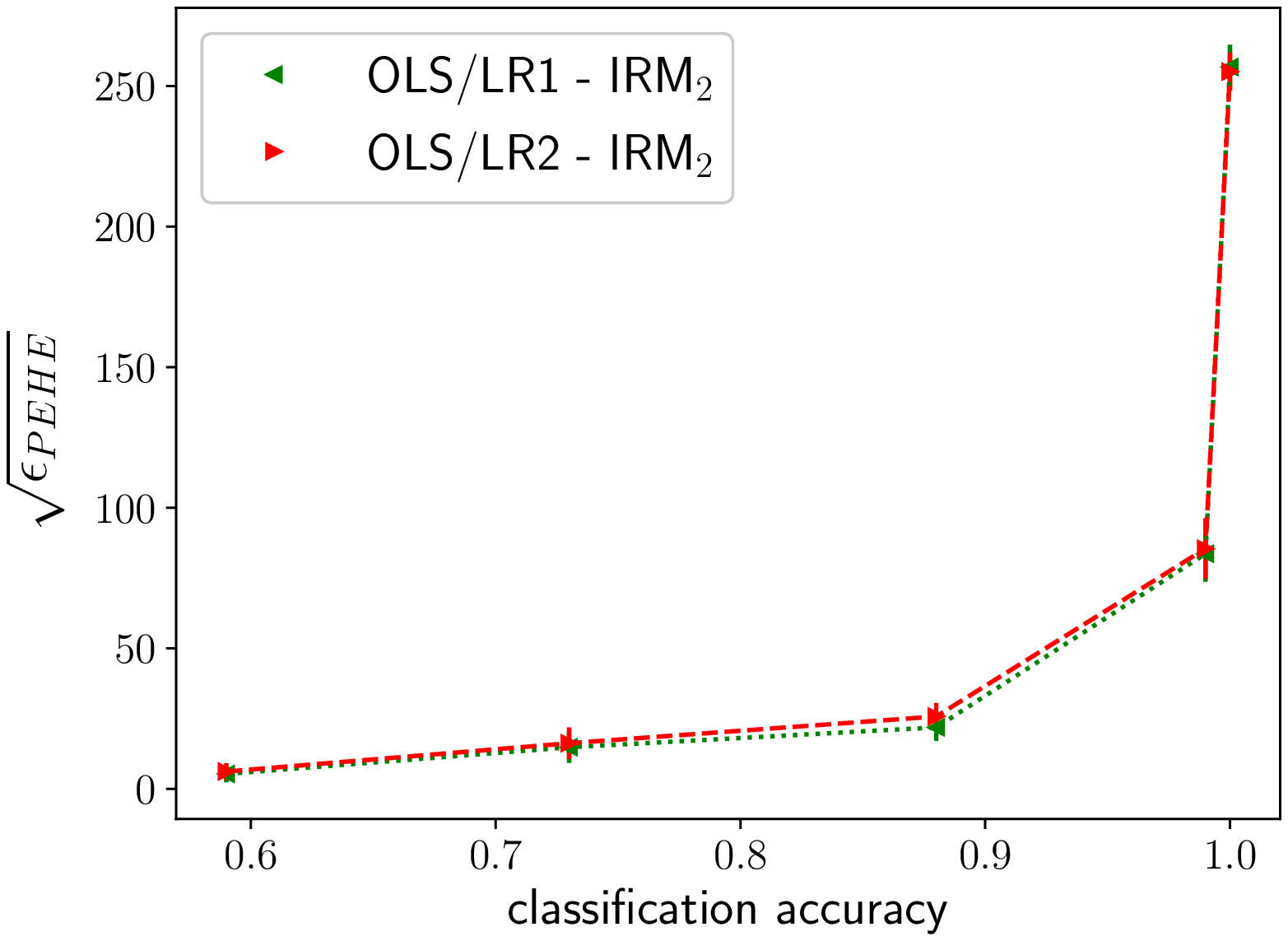}}\caption{$\sqrt{\pehe}$ (left) and $\sqrt{\pehe}$ difference (right) vs treatment group classification accuracy for model B with quadratic outcomes}
  \label{fig:class_B}
\end{figure}

To generate the covariance matrices in \eqref{modelA}, \eqref{modelB}, we first draw eigenvalues uniformly from $[0,1]$, place them as the diagonal entries of diagonal matrices $\bLambda$, $\bLambda_0$, and $\bLambda_1$, and re-scale them to sum to 1. The entries of $\bLambda$ and $\bLambda_0$ are placed in increasing order and the entries of $\bLambda_1$ are in decreasing order. We then let $\bSigma$ = $\bQ_A \bLambda \bQ_A$ in \eqref{modelA}, $\bSigma_0$ = $\bQ_B \bLambda_0 \bQ_B$ in \eqref{modelB} and $\bSigma_1$ = $\bQ_B \bLambda_1 \bQ_B$ in \eqref{modelB} for two orthonormal eigenvector matrices $\bQ_A$ and $\bQ_B$ with entries drawn from $\cN(0,1)$. 
We choose the coefficients $c_0$, $c_1$ in \eqref{linear}, \eqref{quadratic}, the entries of the vectors $b_0$, $b_1$ in \eqref{linear}, \eqref{quadratic}, and the entries of the matrices $A_0$, $A_1$ in \eqref{quadratic} from the uniform distribution over $[0,1]$\footnote{In the version of this paper presented at ICASSP 2021, the figures for model A were generated by choosing these coefficients from the uniform distribution over $[-1,1]$ and the figures for model B were generated by choosing these coefficients from the uniform distribution over $[0,1]$. In the current version, we generate the figures for both model A and B by choosing these coefficients from the uniform distribution over $[0,1]$.}. We let $\sigma$ in \eqref{linear}, \eqref{quadratic} be 1.

For the first set of experiments, we quantify the mismatch between \textit{the control group} and \textit{the treatment group} using the \emph{classification accuracy} in distinguishing between the groups, i.e., predicting treatment assignment with features $\rvbx$ as input to the classifier, 
$p(\rvt = 1 | \rvbx)$. With $d = 35$, we vary the $\bmu_0$ and $\bmu_1$ in \eqref{modelA} and \eqref{modelB} (i.e., the length of \centerLine in Fig.\ \ref{fig:illustrative_example}) to vary the separation between \textit{the control group} and \textit{the treatment group} and in-turn vary the classification accuracy. Fig.\ \ref{fig:class_A} shows that as the distributions of \textit{the control group} and \textit{the treatment group}, for model A with quadratic outcomes, become more mismatched i.e., as the classification accuracy increases, the gains of IRM$_2$ over OLS/LR1, and OLS/LR2 start increasing. Fig.\ \ref{fig:class_B} shows the same for model B with quadratic outcomes. We do not show similar plots for the linear outcome generation method because, in the relatively simpler linear setting, the gains of IRM$_2$ are visible only when classification accuracy is very close to one.

\begin{figure}
  \centering
  \subfloat{\includegraphics[scale=0.25]{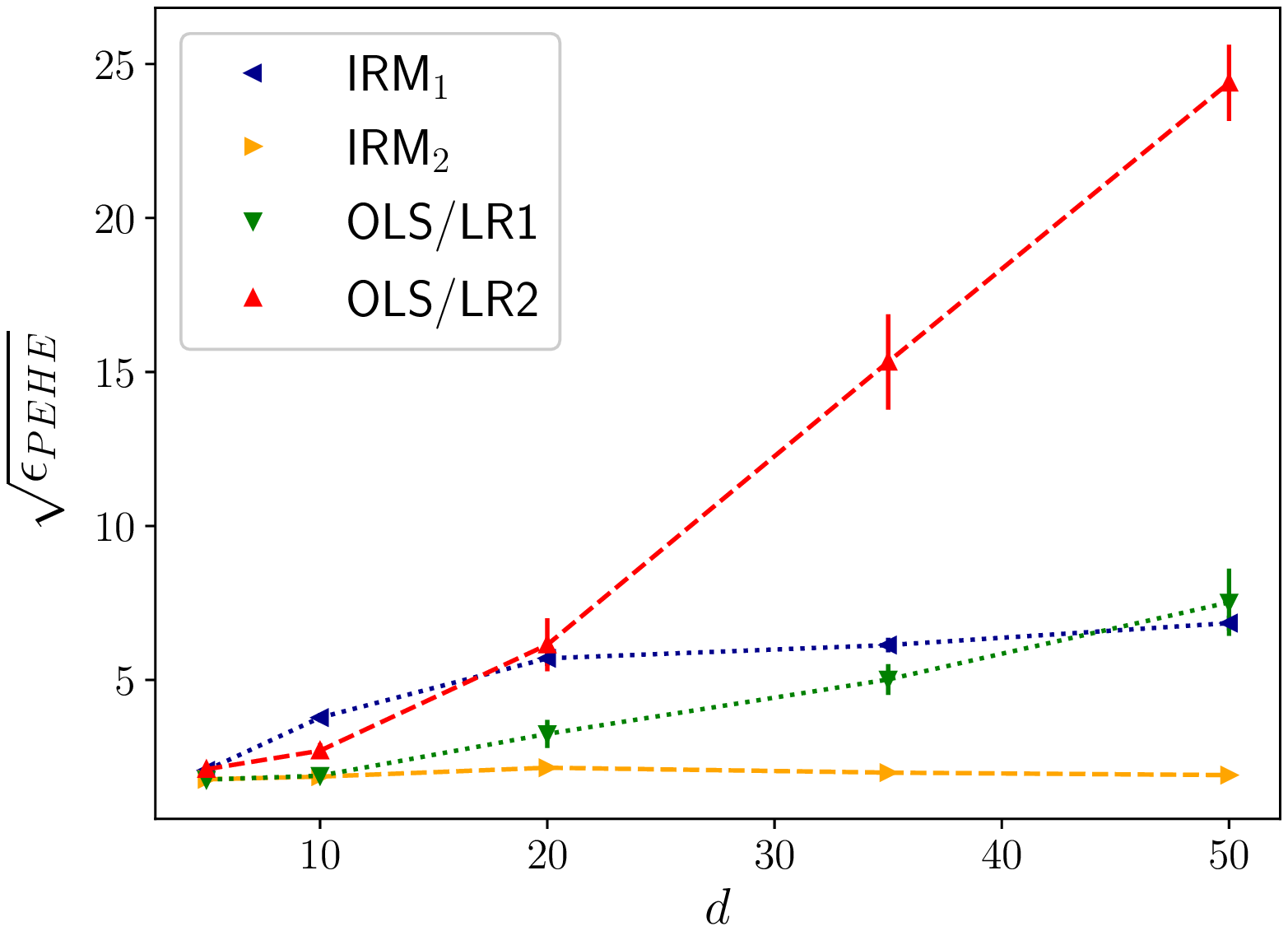}}\quad
  \subfloat{\includegraphics[scale=0.25]{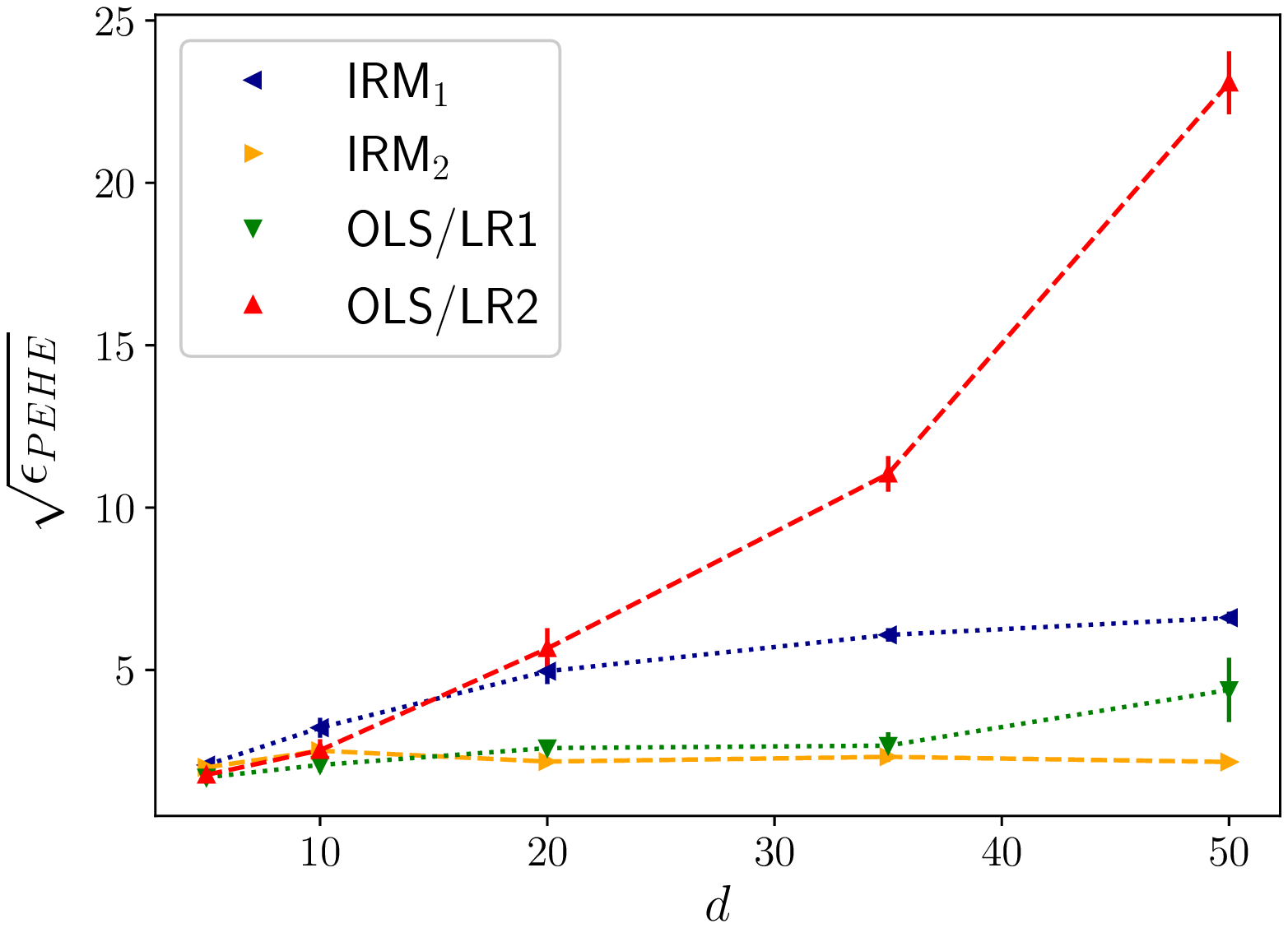}}
  \caption{$\sqrt{\pehe}$ vs $d$ for models A and B with linear outcome}
  \label{fig:linear}
\end{figure}

\begin{figure}
  \centering
  \subfloat{\includegraphics[scale=0.25]{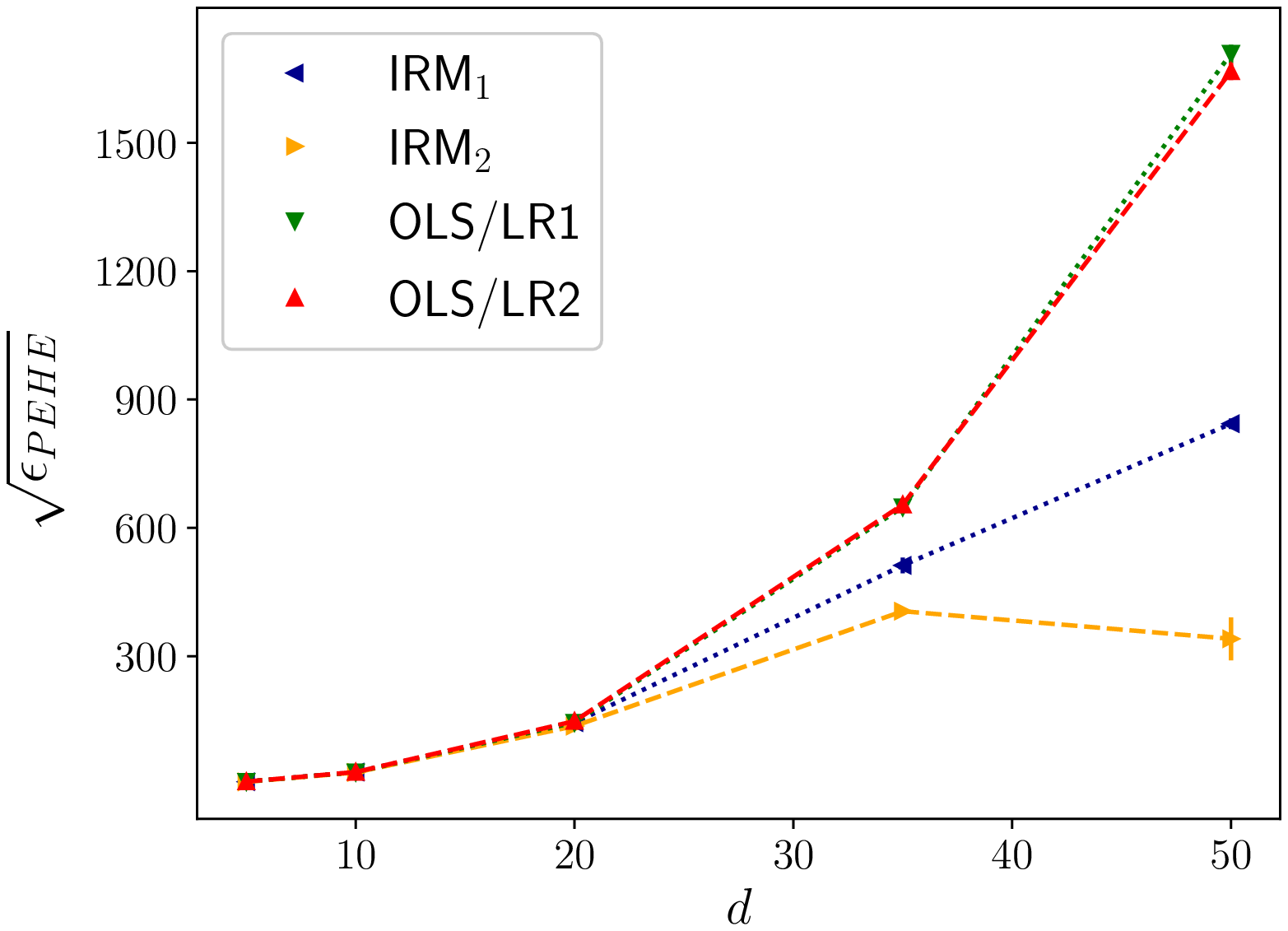}}\quad
  \subfloat{\includegraphics[scale=0.25]{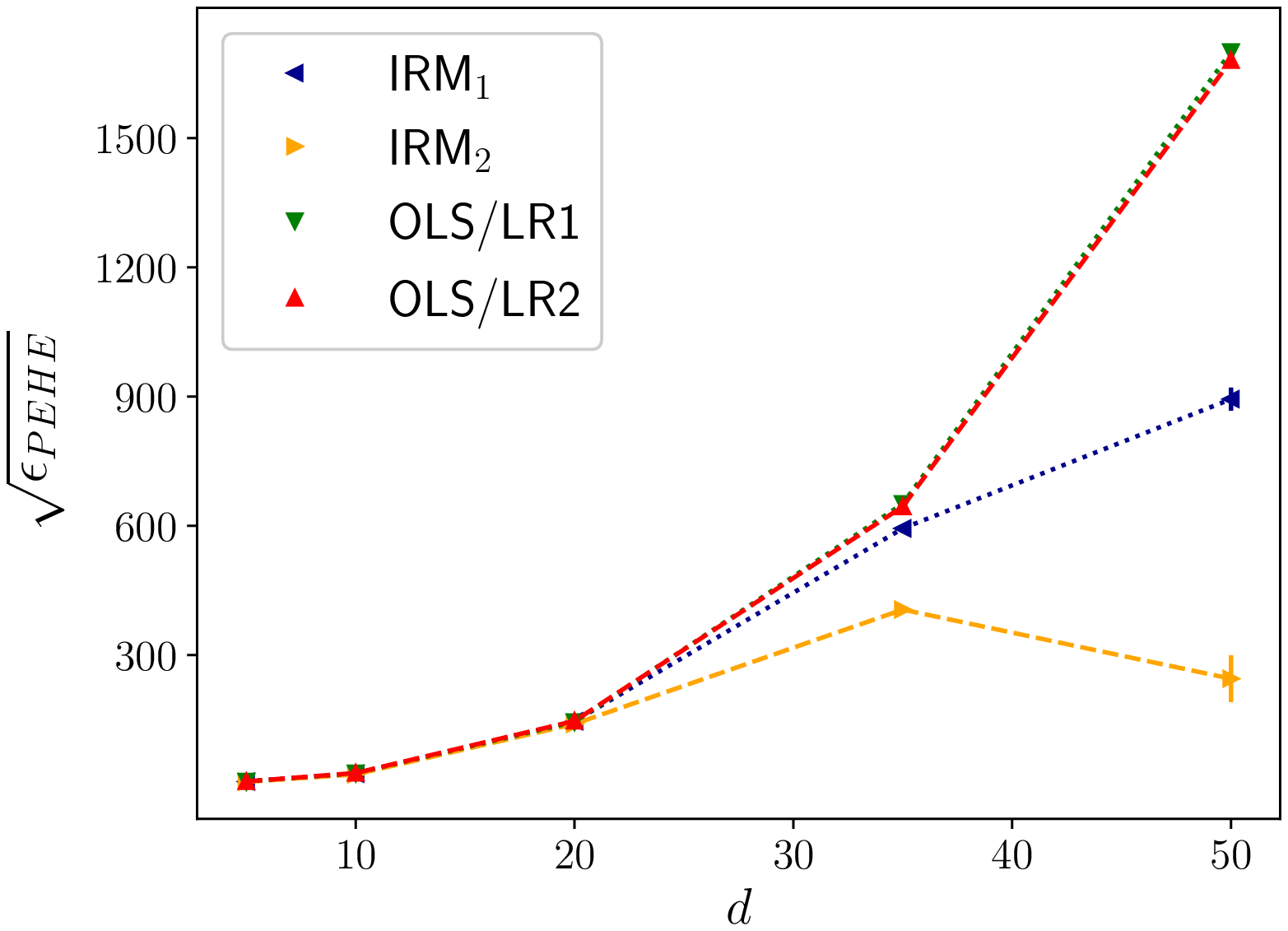}}
  \caption{$\sqrt{\pehe}$ vs $d$ for models A and B with quadratic outcome}
  \label{fig:quadratic}
\end{figure}

For the second set of experiments, for the linear outcome models, we let $\bmu_0$ in \eqref{modelA} to be all -1's and $\bmu_1$ in \eqref{modelB} to be all +1's. Similarly, for the quadratic outcome models, we let $\bmu_0$ in \eqref{modelA} to be all -0.1's and $\bmu_1$ in \eqref{modelB} to be all +0.1's. We vary the dimension $d$ as 5,10,20,35,50 and plot the PEHE for IRM$_2$, IRM$_1$, OLS/LR2, and OLS/LR1 for the linear outcome models in Fig.\ \ref{fig:linear} and for the quadratic outcome models in Fig.\ \ref{fig:quadratic}. For linear models, IRM$_2$ outperforms the other methods in high dimensions and we need a greater mismatch between \textit{the control} and \textit{the treatment} groups i.e., $\bmu$'s to be -1's and +1's to achieve this. For quadratic models, both IRM$_2$ and IRM$_1$ outperform OLS/LR2 and OLS/LR1 even in the regimes with lower mismatch between \textit{the control group} and \textit{the treatment group}, i.e., $\bmu$'s to be -0.1's and +0.1's.

The source code of our implementation is available at - \\
\url{https://github.com/IBM/OoD/tree/master/IRM_ITE}

%% file: tex/data_generation.tex
\subsection{Data Generation}
\label{sec_generative_model}
In our data generation mechanism, we first generate the treatment, followed by the features conditional on the treatment, and finally the outcomes conditional on the treatment and the features.
\begin{enumerate}[leftmargin=*]
        \item {\bf Treatment generation:} Treatment assignments are drawn from a Bernoulli distribution with mean 0.5 i.e., $\rvt \sim$ Bernoulli$(0.5)$.
        \item {\bf Feature generation:} Given the treatment assignment, we consider two feature generation models.\\
                 \begin{align}\label{modelA}
                \rvbx | \rvt= t & \sim \cN(\bmu_t, \bSigma)
                \end{align}
                 $\bullet$ {\bf Model B:} In the second, the features for different groups (\textit{the control} and \textit{the treatment}) are drawn from different multivariate Gaussian mixture distributions as follows. For $t \in \{0,1\}$,
                \begin{align}\label{modelB}
                \rvbx | \rvt = t & \sim 0.5 \times \cN(\bmu_t, \bSigma_0) + 0.5 \times  \cN(\bmu_t, \bSigma_1)
                \end{align}
        \item {\bf Outcome generation}: Given the features and the treatment assignment, we consider two outcome generation methods.\\
                $\bullet$ {\bf Linear:} In the first,  outcomes for different groups are drawn from Gaussian distributions with means given by group-dependent linear functions of the features. For $t \in \{0,1\}$,
                \begin{align}\label{linear}
                y_t | \rvbx = \svbx, \rvt = t  &\sim \cN(\svbx^Tb_t + c_t, \sigma^2)
                 \end{align}
                 $\bullet$ {\bf Quadratic:} In the second,  outcomes for different groups are drawn from Gaussian distributions with means given by group-dependent quadratic functions of the features. For $t \in \{0,1\}$,
                \begin{align}\label{quadratic}
                y_t | \rvbx = \svbx, \rvt = t  &\sim \cN(\svbx^TA_t\svbx+ \svbx^Tb_t + c_t, \sigma^2)   
                \end{align}
        Given the treatment assignment and the potential outcomes, the factual outcomes are: $y_f = t \times y_1 + (1-t) \times y_0$. We know the true potential outcomes and therefore the ITE using \eqref{ite}.
\end{enumerate}

%% file: tex/conclusion.tex
\vspace{-1mm}
\section{Conclusion}
\vspace{-0.5mm}
We have developed an approach for making ITE estimation robust to treatment assignment bias using the domain generalization framework of IRM. We use IRM base-learners inside the S-learner and T-learner frameworks for ITE estimation. In contrast to the typical setting for IRM, we do not require datasets coming from different domains, but create diverse partitions as part of the inference method. We see from our experiments that in scenarios with treatment assignment bias, IRM captures fewer biases compared to OLS. As the treatment assignment bias increases, the reduction in the bias of IRM becomes more prominent.